\documentclass[10pt,twocolumn,letterpaper]{article}

\usepackage{wacv}
\usepackage{times}
\usepackage{epsfig}
\usepackage{graphicx}
\usepackage{amsmath}
\usepackage{amssymb}
\usepackage{booktabs}
\usepackage{float}
\usepackage{dblfloatfix}
\usepackage{multirow}
\usepackage[accsupp]{axessibility}

\def\wacvPaperID{16} % Enter the WACV Paper ID here

\wacvfinalcopy % *** Uncomment this line for the final submission

\ifwacvfinal
\def\assignedStartPage{1} % *** Enter the assigned starting page number (instead of 9876)
\fi

\ifwacvfinal
\usepackage[breaklinks=true,bookmarks=false]{hyperref}
\else
\usepackage[pagebackref=true,breaklinks=true,colorlinks,bookmarks=false]{hyperref}
\fi

\ifwacvfinal
\else
\fi

\begin{document}
%%%%%%%%% TITLE
\title{OTB-morph: One-Time Biometrics via Morphing applied to Face Templates}
\author{Mahdi Ghafourian, Julian Fierrez, Ruben Vera-Rodriguez, Ignacio Serna, Aythami Morales\\
Biometrics and Data Pattern Analytics, BiDA-Lab, Universidad Autonoma de Madrid, Spain\\
{\tt\small (mahdi.ghafourian,julian.fierrez,ruben.vera,ignacio.serna,aythami.morales)@uam.es}
%\and
}

\maketitle
%\thispagestyle{empty}

%%%%%%%%% ABSTRACT
\begin{abstract}
   Cancelable biometrics refers to a group of techniques in which the biometric inputs are transformed intentionally using a key before processing or storage. This transformation is repeatable enabling subsequent biometric comparisons. This paper introduces a new scheme for cancelable biometrics aimed at protecting the templates against potential attacks, applicable to any biometric-based recognition system. Our proposed scheme is based on time-varying keys obtained from morphing random biometric information. An experimental implementation of the proposed scheme is given for face biometrics. The results confirm that the proposed approach is able to withstand against leakage attacks while improving the recognition performance.
\end{abstract}

%%%%%%%%% BODY TEXT
\section{Introduction}

The advantages of biometric recognition in authentication systems against conventional methods such as password or smart cards have attracted much attention to this field. However, the widespread usage of biometrics has raised serious security and privacy concerns \cite{2021_EncCrypto_BioSec_Fierrez,Galbally2007_Vulnerabilities}. In addition, standard cryptographic approaches cannot be directly applied to solve these security threats due to the variable and noisy nature of biometrics \cite{Freire2007_ICB}. Therefore, a new class of methods called Biometric Template Protection (BTP) emerged as a remedy \cite{2017_Access_HEmultiDTW_Marta,2017_PR_multiBtpHE_marta,7192825}.  Biometric template protection refers to a set of techniques to preserve the security and privacy of the acquired biometric data. The main goal is to generate a protected biometric reference guaranteeing: noninvertibility (irreversibility), revocability (renewability), and unlinkability (nonlinkability); without degrading the recognition performance. BTP methods are commonly divided into three categories \cite{1patel2015cancelable}: cancelable biometrics \cite{10maiorana2010cancelable}, biometric cryptosystems \cite{1299169}, and biometrics in encrypted domains \cite{2017_PR_multiBtpHE_marta}.

Cancelable biometrics refers to a group of biometric template protection techniques with the primary aim of improving template security and privacy by obscuring the original feature using an irreversible but repeatable transformation such that the recognition still can be performed only in the transformed domain. These methods should maintain four characteristics: \textit{Diversity}, \textit{Revocability}, \textit{Non-invertibility}, and \textit{Recognition performance}. During enrollment, biometric features are extracted upon presentation, then the corresponding cancelable biometric technique is applied to these features (mostly by using auxiliary data) and finally the result (transformed template) is stored in a template database (server). During verification, the transformed template of the presented biometrics is obtained similar to the enrollment phase by applying the previously stored or known auxiliary data. Lastly, the matching takes place between the generated cancelable template at the verification phase and the one stored at the enrollment phase called reference. A general taxonomy of all cancelable biometrics methods containing six major categories has been proposed recently in \cite{2kumar2020cancelable}. 

In the present paper we apply the concepts behind one-time pad \cite{doi:10.1080/01611194.2011.583711} to derive one-time biometrics, in a kind of cancelable biometrics. The core elements of our proposed scheme are: (i) to use as time-variant keys biometric data generated randomly with natural appearance \cite{2020_JSTSP_GANprintR_Neves}, (ii) combining these keys (random biometrics) with real input biometric data using image/signal morphing techniques \cite{18scherhag2019face}, and (iii) keeping track of the key/template variations in time in a specific secure exchange protocol to enable biometric comparisons while protecting against potential attacks.

The rest of this paper is organized as follows: Sect. 2 summarizes related work in cancelable biometrics. Sect. 3 describes the attack framework we have considered for evaluating the security improvement that our proposed method can provide. Sect. 4 describes our proposed scheme for one-time biometrics. Experimental results applying the proposed concepts to face biometrics are reported in Sect. 5. Sect. 6 concludes the paper.

\section{Related Works in Cancelable Biometrics}

Over the past two decades, many cancelable biometrics research has been carried out due to the increasing usage of biometric-based authentication. Here we review some early and noticeable attempts in this area. 

\begin{figure*}[tbh]
 \centering 
 \includegraphics[trim={3cm 4cm 2cm 2cm},clip,width=170mm,scale=0.5]{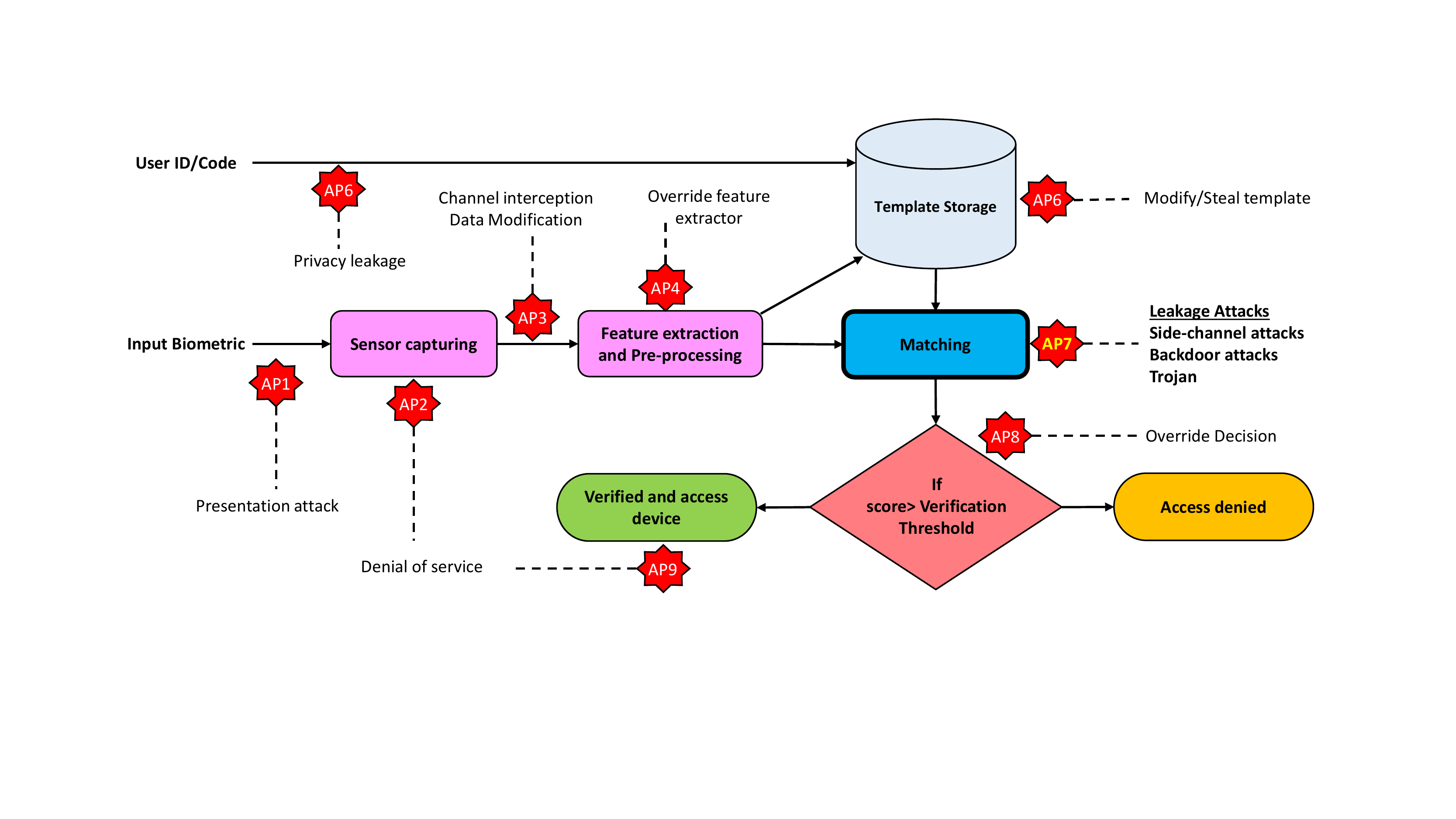}
 %trim={<left> <lower> <right> <upper>}
  \caption{Attack points in a generic biometric system.}
  \label{fig:AttackPoints}
\end{figure*}

The concept of cancelable biometrics was first introduced in \cite{3ratha2001enhancing} to enhance the security and privacy in biometric-based authentication systems. Among early noticeable attempts, Jin et al. \cite{4jin2004biohashing} proposed a Random projection-based technique called BioHashing. This method projects biometric features to a random space by taking the inner product between a tokenised pseudo-random number and the user fingerprint. In 2005, Ang et al. \cite{5ang2005cancelable} proposed a key-dependent cancelable template where a geometric transformation was applied to features extracted from a fingerprint so as to protect minutiae templates. In 2006, Chin et al. \cite{6chin2006high} presented a work securing iris features coined as S-Iris encoding. To this end, they iterated inner products between secret pseudo-random numbers and the iris features. In 2007, the first alignment-free cancelable biometrics was introduced by Lee et al. \cite{7lee2007alignment}. They protected fingerprint templates by extracting rotational and translational invariant features from each minutia. Later that year, Ratha et al. \cite{8ratha2007generating} suggested three different methods (Cartesian, Polar, and Surface Folding) to transform minutia positions extracted from a fingerprint image. These transformations were aimed at distorting original biometrics and offering noninvertibility and revokability. However, soon after Quan et al. \cite{9quan2008cracking} showed that most of the transformed minutia in \cite{8ratha2007generating} could be exactly inversed. 

More recently Maiorana et al. \cite{10maiorana2010cancelable} proposed a convolution-based noninvertible transformation named BioConvolving, which can be applied to any sequence-based biometric. They practiced their approach on online signature biometrics and its security relies on the difficulty of solving a blind deconvolution problem. Same year, Ouda et al. \cite{11ouda2010tokenless} proposed a cancelable biometric scheme for protecting Iris-Codes. Their method extracts consistent bits from Iris-Codes and further encode them using a random encoding process referred to as BioEncoding. Same year, another research \cite{12pillai2010sectored} generated cancelable iris biometrics using sectored random projections. This method mitigates the performance degradation due to eyelids and eyelashes. In 2012, Ferrara et al. \cite{13ferrara2012noninvertible} provided noninvertibility based on dimensionality reduction and binarization to protect Minutia-Cylinder-Code, which is a local minutia representation. Later, Gomez-Barrero et al. \cite{2016_InformationSciences_Marta,marta14FaceBF,rathgeb15IWBF_IrisFaceBF} proposed an alignment-free cancelable iris template based on Bloom filters. They argued that successive mapping of parts of a binary biometric template to a Bloom filter represent a noninvertible transformation. Chin et al. \cite{15chin2014integrated} proposed another template protection technique in 2014 by fusing fingerprint and palmprint at the feature level on the basis of user-specific keys. Three years later, Lai et al. \cite{16lai2017cancellable} introduced a cancelable iris template generation method coined as Indexing-First-One (IFO) hashing. The method is inspired from Min-hashing and further strengthening by using modulo threshold function and P-order Hadamard product. Finally, Sadhya and Raman \cite{17sadhya2019generation} generated a cancelable iris template using randomized bit sampling. Their method (LSC) is functionally based on the notion of Locality Sensitive Hashing (LSH) in which two items that are relatively close to each other, hash into the same location \cite{Freire2007_ICB}.

%-------------------------------------------------------------------------
\section{Adversary Model}

\label{sec:adversary}

%\begin{figure*}[tbh]
%\centering
%  \includegraphics[width=150mm,scale=0.5]{Attack points}
%  \caption{Attack points in a generic biometric system.}
%  \label{fig:AttackPoints}
%\end{figure*}

\begin{figure*}[tbh]
 \centering 
 \includegraphics[trim={3cm 4cm 2cm 2cm},clip,width=170mm,scale=0.5]{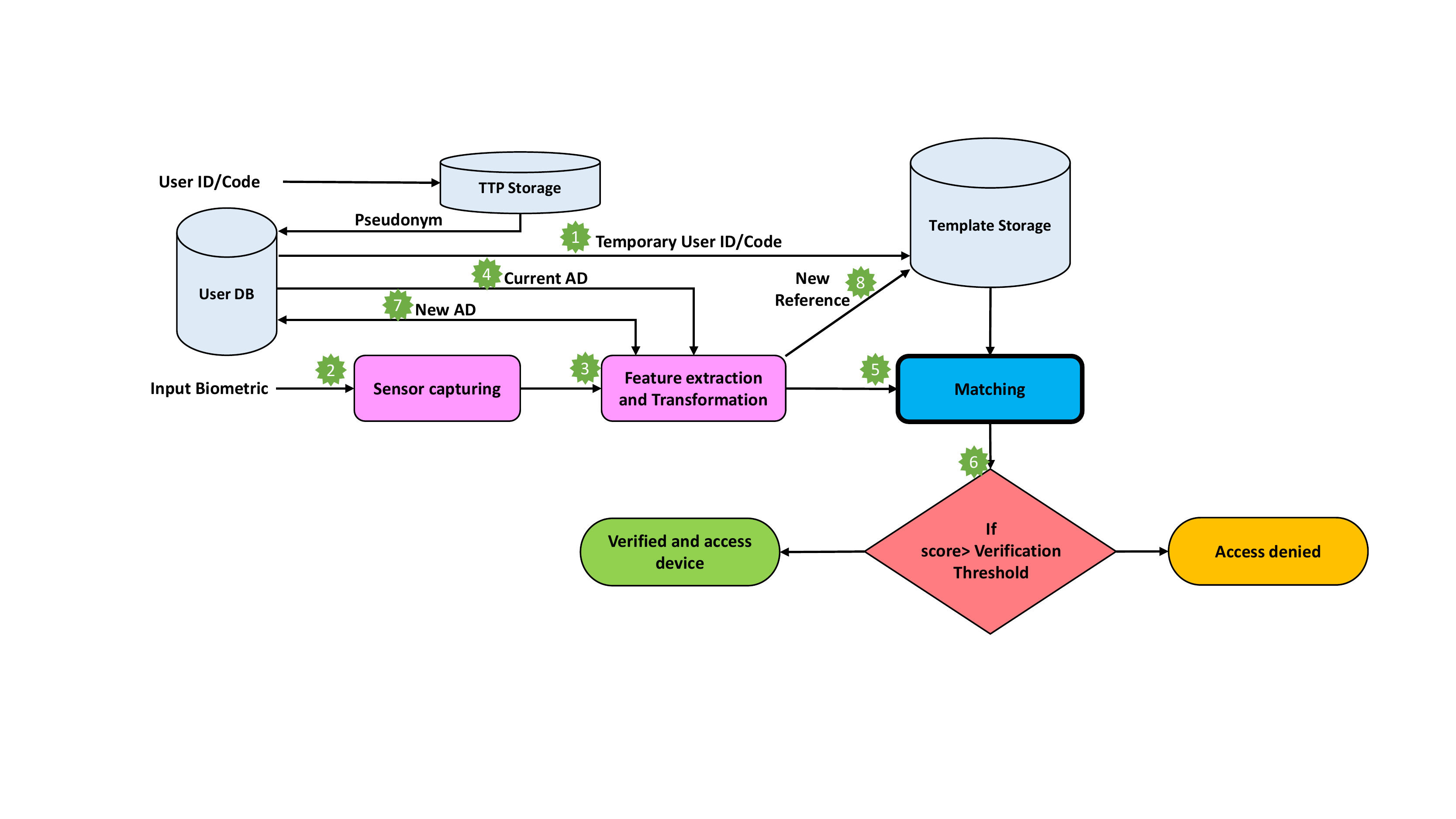}
 %trim={<left> <lower> <right> <upper>}
  \caption{Architecture of the proposed One-Time Biometrics scheme (OTB-morph).}
  \label{fig:Architecture}
\end{figure*}

Biometric systems can be the target for an attacker to conduct malicious activities, including impersonation. The possible attack points are positioned in a generic biometric system in Figure~\ref{fig:AttackPoints} \cite{2021_EncCrypto_BioSec_Fierrez,Galbally2007_Vulnerabilities,19jaswal2021ai}. This paper is focused on addressing three challenges: (i) privacy leakages at attack point AP6, (ii) injection attack at AP4, and (iii) leakage threats at the AP7. In particular, we assumed:

\begin{itemize}
 \item The attacker is able to eavesdrop the communication channel from AP6 where genuine users request verification.
  \item The similarity score of biometric templates at the matching phase is leaked to the attacker through any wide-range means of leakage attacks such as backdoors, trojans, side-channel attacks \cite{GALBALLY2020101902,2009_GalballyBioID}, etc. 
  \item The attacker is able to get the similarity score between an arbitrary biometric input and the feature reference of victims from AP7 for some verification sessions, not necessarily consecutive.
  \item The attacker possesses the knowledge of the underlying model with which the protected template (victim's reference) is generated from the input biometric data (i.e., the biometric feature extractor).
  \item The attacker is able to get at least one biometric input of the victim.
  \item The attacker is able to override the feature extractor and can inject his biometric features in AP4. 
\end{itemize}

Using this leaked score or the obtained biometric input, the attacker can maximize the similarity of his arbitrary input biometric compared to the victim's reference by iterative optimization, e.g., deep leakage from gradient \cite{zhu2020deep}, hill-climbing \cite{Galbally_2009PR,barrero13PRLmultimodalAttack,barrero12ICB}. 

%\begin{equation} \label{eq1}
%    \frac{\partial J}{\partial w}=\frac{1}{m}X(A-Y)^{T}
%\end{equation}
%\begin{equation} \label{eq2}
%    \frac{\partial J}{\partial b}=\frac{1}{m}\sum_{i=1}^{m} (a^{i}-y^{i})
%\end{equation}
%
%Where $A=(a^1,a^2,…,a^m)$ is the output of applying activation function on the network input function, ${X=(x^1,x^2,…,x^m)}\in\mathbb{R}^{n_x}$ is input feature vector, $Y=(y^1,y^2,…,y^m)$ is the target similarity score (Ground truth), and m is the number of training examples in the attacker's network.

\section{Proposed Scheme: OTB-morph}

%\begin{figure*}[htp]
%\centering
%  \includegraphics[width=150mm,scale=0.5]{Architecture}
%  \caption{Architecture of the proposed One-Time Biometrics scheme (OTB-morph).}
%  \label{fig:Architecture}
%\end{figure*}
The aim of the proposed scheme is to address both privacy leakages at attack point 6 (AP6, see Figure~\ref{fig:AttackPoints}) and leakage attacks at attack point 7 (AP7). The block diagram of the proposed scheme is shown in Figure~\ref{fig:Architecture}. 

\begin{figure*}[tbh]
 \centering 
 \includegraphics[trim={0 0cm 0 0cm},clip,width=170mm,scale=0.5]{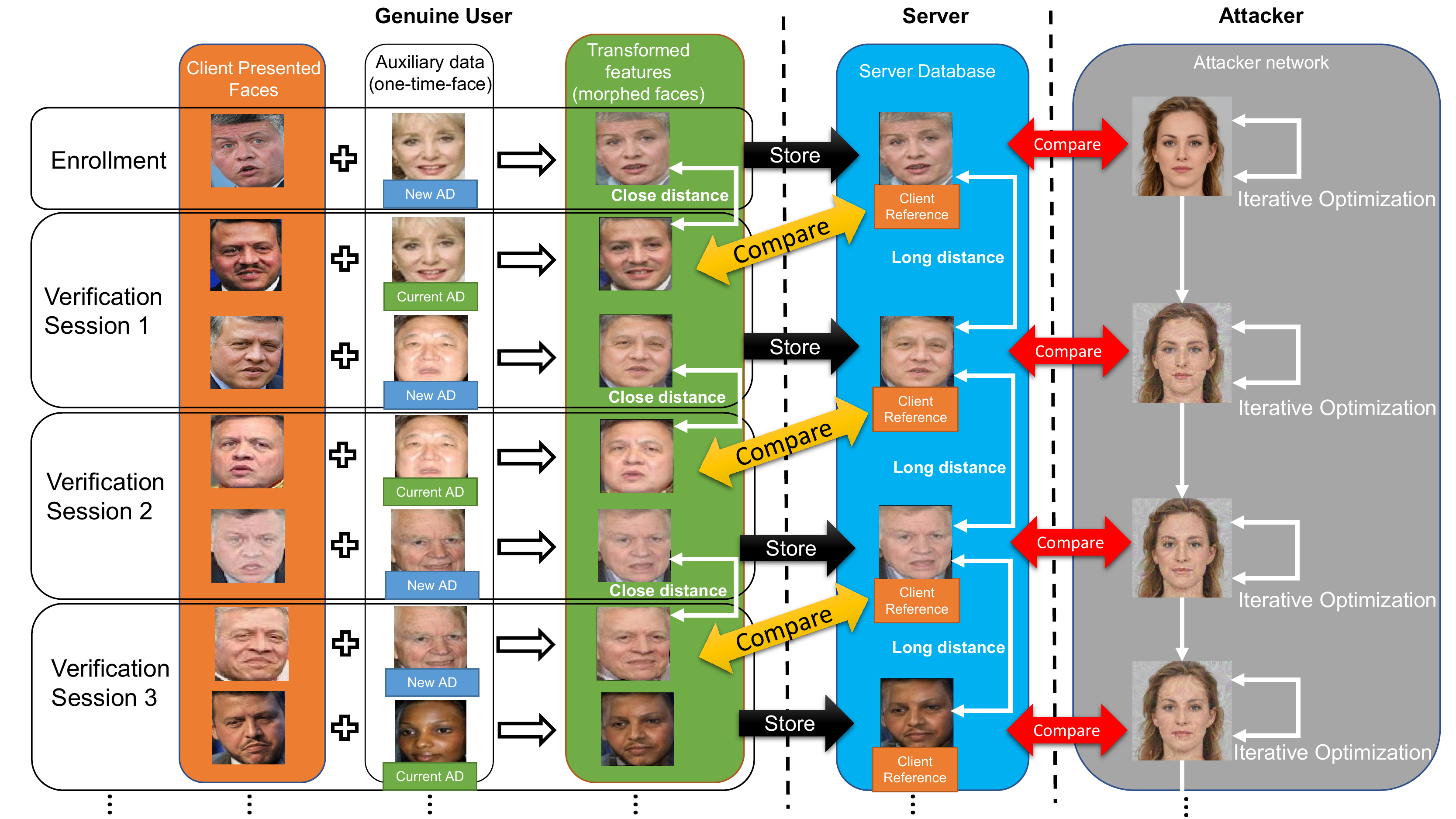}
 %trim={<left> <lower> <right> <upper>}
 \caption{Visual examples of the process of the proposed One-Time Biometrics via Morphing (OTB-morph) for enrollment and various verification sessions (genuine users and attackers).}
 \label{fig:process}
\end{figure*}

There are three parties involved during biometric verification. A \textit{Client} who wants to be verified in a \textit{Server} using a temporary identity that has been assigned to him by a \textit{Trusted Third Party (TTP)}. It is assumed that enrollment phases in both server and TTP are already accomplished and the corresponding Auxiliary Data (AD) and Pseudonyms are stored on a secure element in the client’s device or his smartcard (note that the complete process of the proposed method is explained in detail later with an example in face biometrics). In this regard, the client starts the verification session by sending his request to the sever using one of his stored pseudonyms (num 1). Pseudonyms are temporary identities that have been assigned to the client prior by the Trusted Third Party (TTP). We refer readers to \cite{20ghafoorian2020anonymous} to study the pseudonym architecture that we are using in this paper. Upon receiving the answer from the server, the client presents his biometric to the input sensor (num 2) and the extracted feature will be transformed to a cancelable biometric template (num 3) using the current AD (num 4) that he has stored on his device/smartcard from the enrolment process. In the next step, the produced cancelable biometric template is sent to the server domain to be compared in the biometric matcher with the feature reference of the client (num 5). Depending on the verification threshold, access is granted or denied (num 6). Generally, most cancelable biometrics techniques need Auxiliary Data (AD) to compute the transformation of biometric features. This AD can be a password, a random number, etc., and it is usually permanent until a leakage on the respective cancelable template is reported. In our proposed method, this auxiliary data are random biometrics (e.g., GAN-generated synthetic faces \cite{2020_JSTSP_GANprintR_Neves}, LSTM-generated synthetic handwriting \cite{2021_AAAI_DeepWriteSYN_Tolosana}, etc.), sent to the user inside the pseudonym sets managed by the TTP. When the matching is successful, we propose to re-enroll by picking a new random biometric (AD) (num 7) and combine it to the already extracted feature. The resulted cancelable template is stored as new reference (num 8) in the server's database. Finally, the new AD is stored on the client's device replacing the previous one. Here with OTB-morph we propose to combine the random and the input raw biometrics via image- or signal-morphing, depending on the nature of the biometrics at hand.

For the better understanding of readers, the whole operation of the proposed cancelable biometrics scheme including the primary enrollment of a user and three sessions of verification is depicted in Figure~\ref{fig:process}. There exists three parties in this process involving a \textit{Genuine User} who attempts to be authenticated in the \textit{Server} in the presence of an \textit{Attacker} who according to the adversary model is able to maximize the similarity of his face to that of the genuine user. The whole process of this figure is described next.

\subsection{Enrollment}
The genuine user enrolls in the server by presenting his face. Upon this, the system picks a random pseudonym and applies a random face image as auxiliary data to the cancelable method. This face image is an arbitrary face image (real or artificial) which is not repeated in any pseudonym sets before or in the future. Then, face morphing transformation is applied to both face images to generate the protected template. Next, the cancelable template is stored on the server's database as the user biometric reference. Finally, the arbitrary face extracted earlier from the pseudonym set is recorded as the current auxiliary data (current AD) in a secure element at the user's device and the corresponding pseudonym is discarded.

\subsection{Verification Session}

Each verification session consists on two steps, as can be seen in Figure~\ref{fig:Architecture}. 

\textbf{Step 1:} Upon presenting the user's face to request verification, the system restores the previously recorded face from the secure element as current AD and computes the morphing. If the matching score between the user's transformed face and his reference is below the threshold (we use dissimilarity scores = distances), then the user is verified and the system runs the second step. Otherwise, it terminates the session.

\textbf{Step 2:} Upon successful verification, the system captures another face image of the user. Then, it picks a random pseudonym set, extracts the arbitrary image inside it as auxiliary data (called new AD) and carries out the morphing to generate a new protected template. Finally, the system overwrites the current AD with the new AD in the secure element and replaces the previous reference with the new protected template. This step is actually doing a re-enrollment of the user. 

\subsection{Attacker Behaviour}
During this process, the attacker tries to maximize the similarity of his face image by comparing it in multiple iterations either to that of the user's face image locally or taking advantage of the leaked matching score (yellow arrow in Figure~\ref{fig:process}). Assuming that the dissimilarity of the user's reference to his raw trait is above the matching threshold and the user's biometric reference is changed at the end of each verification session, neither of these attacks would be successful in the proposed method. In other words, while the Euclidean distance between two templates, which were morphed using the same auxiliary data is low (below the threshold), the same distance between those that are transformed by different auxiliary data is high (above threshold). These unique characteristics of the proposed method prevents the attacker to maximize his face image and thus impersonate the genuine user.

\section{Experiments}

As indicated before (cf. Section~\ref{sec:adversary}), for our security analyses we assume that the adversary has access to the matching score of victims and he is able to update an arbitrary face image such that the corresponding score (Euclidean Distance in our experiments, therefore dissimilarity score) of it with respect to the victim's reference becomes lower than the verification threshold \cite{Galbally_2009PR}. In other words, the adversary is able to manipulate an arbitrary face image and successfully impersonates a legal client. In order to evaluate the weakness of current cancelable biometrics methods against this kind of leakage attacks, we carried out our experiments comparing four scenarios: (i) face verification without applying any protection method; (ii) face verification protected by applying Gaussian noise as cancelable transformation to probe feature \cite{1patel2015cancelable}; (iii) face verification protected by applying imploding, a cancelable image transformation pulling pixels into the middle of the image \cite{1patel2015cancelable}; and (iv) face verification protected by applying the proposed method OTB-morph. The experiments are conducted on three face datasets: VGGFace2 \cite{21cao2018vggface2}, Casia \cite{22yi2014learning}, and LFW \cite{2018_TIFS_SoftWildAnno_Sosa,24huang2008labeled}.

\subsection{Implementation Details}

The implementation is performed on a pretrained Resnet-50 \cite{23he2016deep}, a CNN model proposed for general image recognition tasks using two groups of datasets. As first group we used VGGFace2 \cite{21cao2018vggface2} and Casia \cite{22yi2014learning} datasets, two face datasets which contain multiple faces of the same individual. The images in these datasets are utilized as probe faces during verification sessions. Regarding the second group, we used LFW \cite{2018_TIFS_SoftWildAnno_Sosa,24huang2008labeled} as the auxiliary data (a random seed) to create morph faces for our proposed OTB-morph scheme. In other words, our method takes two input faces, one from the first group as the probe biometric feature of the subject meant to be protected, and the second input is a randomly chosen face image from the second group to be morphed with the first image. 

\subsubsection{Image Morphing}
\begin{figure}[b]
\begin{center} \includegraphics[width=\linewidth]{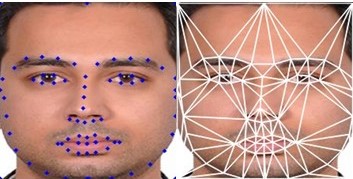}
\end{center}
   \caption{Example of landmark-based morph generation.}
\label{fig:morphed}
\end{figure}

Image morphing is an image processing technique that can transform one image to another image. Applied to face images, morphing can be used to produce artificial faces which resemble the biometric characteristics of at least two input individuals in image and feature domains \cite{18scherhag2019face}. Morphed faces can be generated using various methods from simple image overlaying to Generative Adversarial Networks (GAN). The most popular morphing method is landmark-based, which consists of three steps: (i) determining a correspondence between the two contributing face images; (ii) warping, which means distorting both features such that the corresponding facial elements (e.g., eye, nose, mouth) are geometrically aligned; and (iii) blending, which refers to the process of merging the color values of wrapped images. In our experiments, we use landmark-based morphing as transformation function for our proposed cancelable biometrics method. Our morphing implementation is based on Dlib for landmark detection \cite{18scherhag2019face} and OpenCV for image processing \cite{26scherhag2020morphing}, and results in facial landmarks as shown in Figure~\ref{fig:morphed}. 

The landmarks locations obtained from both face images are warped by averaging the pixel positions. After moving the pixels we apply image warping based on Delaunay triangulation \cite{25venkatesh2021face}, as shown in Figure \ref{fig:morphed} right. Our morphing method has a parameter $\alpha$ between 0 and 1 that trades-off the contribution of each input image: smaller $\alpha$ generates an output more similar to the first contributed face image (probe face in our case), and higher $\alpha$ results in a morphed face more alike to the second contributed face image (random face). In these experiments we selected $\alpha=0.5$ to keep the trade-off.

\subsection{Performance and Security Metrics}

We use the Equal Error Rate (EER) to evaluate and compare the verification performance of our proposed method with other scenarios. EER is the point where the False Acceptance Rate (FAR) and False Rejection Rate (FRR) are equal, where FAR is the percent of unauthorized users (random impostors\footnote{This kind of impostors are different to the attackers considered in Section~\ref{sec:adversary}, who have much more information to attack the system compared to a random impostor that just tries to illegally access the system by using his own face input and no other methods to improve the attack success.}) incorrectly verified as a valid user (genuine) while FRR is the percent of incorrectly rejected valid users. The evaluation metric EER describes the overall accuracy of a biometric system. In general, the lower the EER value, the higher the accuracy of the biometric system. 

Regarding security evaluation, the vulnerability of the compared cancelable biometrics schemes under the considered Adversary Model (cf. Section~\ref{sec:adversary}) is analyzed looking at the capability of the attacker to minimize the dissimilarity score of his arbitrary face image by iterative optimization exploiting the leaked matching score. More specifically, we measure the Attack Success Rate (ASR) to assess and compare the vulnerability of all experimental scenarios \cite{Galbally_2009PR}. 

\subsection{Results}

%[TODO, please revise this full section as indicated, the information here is not clear, just a draft to give you ideas on how to present and explain your results. YOU HAVE TO READ THE FIGURES AND TABLES IN THE WAY YOU WANT THE READER TO READ THEM, INCLUDING DESCRIPTION OF INDIVIDUAL PLOTS, AXES, KEY PERFORMANCE NUMBERS, ETC.]%

\begin{figure*}[tbh]
 \centering 
 \includegraphics[trim={0cm 0cm 0 0cm},clip,width=175mm,scale=0.5]{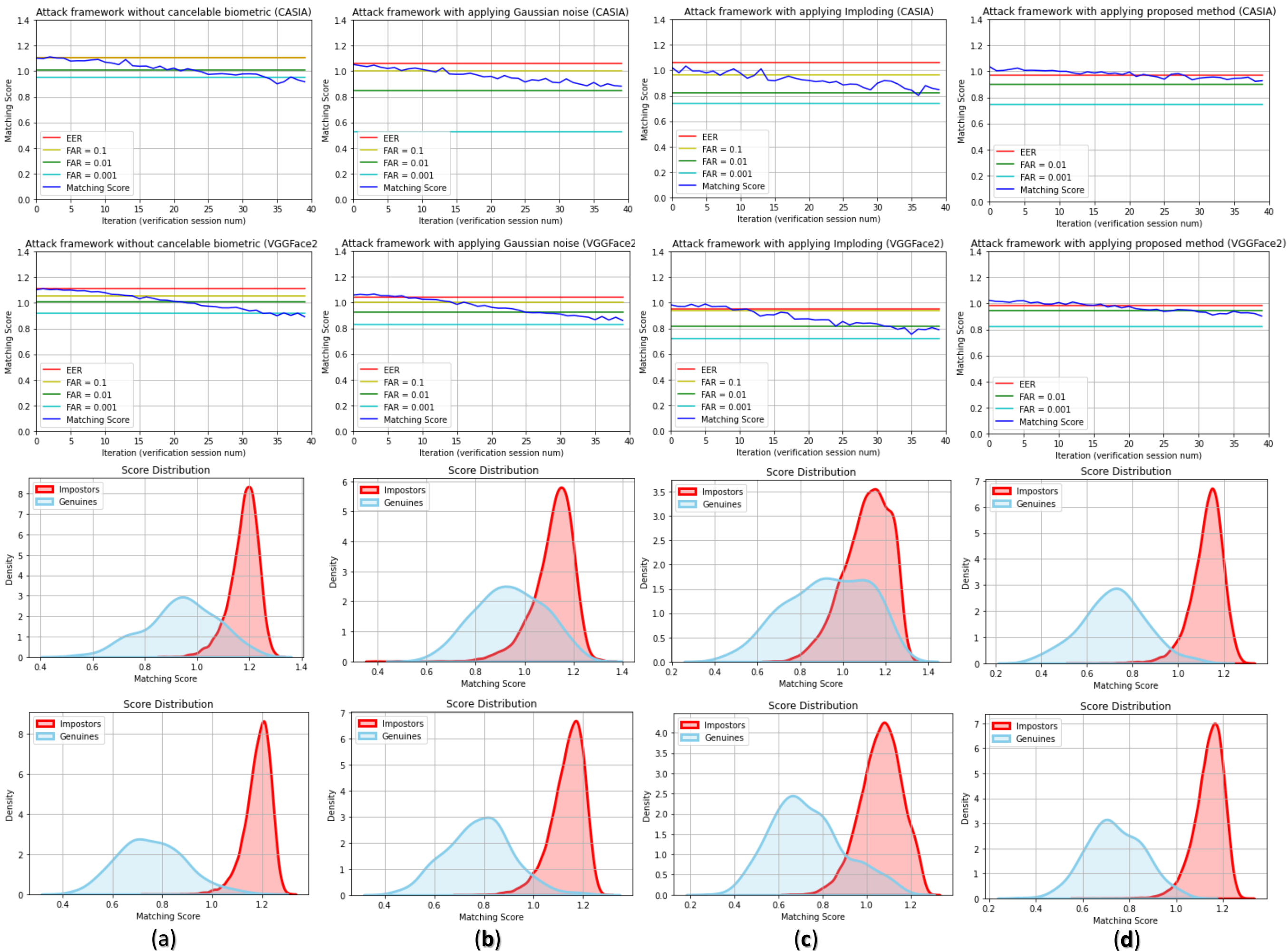}
 %trim={<left> <lower> <right> <upper>}
  \caption{Comparison of practiced scenarios: column (a) without applying cancelable biometrics; column (b) applying Gaussian noise; column (c) applying imploding; and column (d) applying the proposed OTB-morph scheme. First row: attacking matching (dissimilarity) score evolution on CASIA dataset (positioned on top of decision thresholds at EER and various FAR). Second row: idem on VGGFace2 dataset. Third row: Genuine and random impostor distributions of the four considered cancelable biometrics approaches on CASIA dataset corresponding to different columns. Forth row: idem on VGGFace2 dataset.}
  \label{fig:results}
\end{figure*}

The main results of our experiments are shown in Figure~\ref{fig:results}. The figure is comprised of four columns, each of them shows the four scenarios considered: (a) not applying  any  protection  method;  (b)  applying Gaussian noise; (c) applying imploding;  and (d) our proposed method OTB-morph. The figure also comprises four rows, the first one shows the attacking matching (dissimilarity) score evolution on CASIA dataset. The second row shows the same as the first row, but for VGGFace2 dataset. The last two rows show the score distributions obtained for the four scenarios considered with respect to CASIA and VGGFace2 datasets, respectively. In each plot of the first two rows in the vertical axis we can see multiple horizontal lines representing the decision threshold location at EER point and various FAR points (see the figure legends). Additionally, each plot represents the time evolution of the attacking score in 40 consecutive iterations (from left to right in each plot). 

Focusing on the first row for CASIA dataset, the column (a) without cancelable biometrics shows that the attacker matching score goes below the acceptance threshold (a little above 0.9) even for a high security threshold (FAR=0.001). For the next two columns, cancelable biometric with applying Gaussian noise and imploding respectively, the matching score falls in similar values just a little above FAR=0.01. This is not happening in the proposed OTB-morph (column (d)), where the attacking score after 40 iterations goes below the threshold corresponding to the EER, but not below thresholds for FAR $<$ 0.01. Same trends are seen for the case of VGGFace2 (second row) although the scores go below the threshold for FAR $<$ 0.01 in this case. Considering the first two rows, the most apparent evolution that can be understood is the falling rate of attacker matching score. While for the first three columns it decreases steadily to a low Euclidean distance (around 0.8), this pace is far slower for the proposed method, keeping the attacker matching score above 0.9 on both CASIA and VGGFace2 cases. If we focus now on the third row, it can be seen that the overlapping area of the impostor and genuine score distributions for our proposed OTB-morph is much smaller compared to the other three cases. With regard to the score distributions for VGGFace2 (last row), while the performance drop is not as severe as the third row, the performance of the proposed method is better compared to the other scenarios.

Additionally, in Table~\ref{tab:results} we report both Equal Error Rate (EER) and False Rejection Rate (FRR) values (verification performance against random impostors), as well as Attack Success Rates (ASR) against the attackers described in Section~\ref{sec:adversary}, for FAR=$\{0.1, 0.01, 0.001\}$. In that table we can see that the smallest EER and FRR values are obtained by the proposed approach (scenario iv) whereas the highest values (worst performance) is overall reported on imploding for both CASIA and VGFace2. On the other hand, the first scenario (unprotected biometric system) has the highest Attack Success Rate in both cases. Out of the four scenarios, although Gaussian noise (scenario ii) did the worst at FAR = 0.01, 0.001 with corresponding FRR=$74.2\%$, $100\%$ respectively, reported FRR results for imploding is worse than other scenarios in both datasets generally. Conversely, the proposed method acquired the best performance with FRR= $1.93\%$ and $0.16\%$ at FAR=$0.1$ on CASIA and VGGFace2 respectively. It is worth to mention that in the proposed method the EER is higher than the FRR at FAR=$0.1$.

\begin{table*}[tbh]
\centering
\caption{Comparison of performance and security of the proposed method (scenario iv) with other scenarios.}
\label{tab:results}
\resizebox{\textwidth}{!}{%
\begin{tabular}{|c|cccc|cccc|} 
\hline
\multirow{3}{*}{Scenario} & \multicolumn{4}{c|}{CASIA \cite{22yi2014learning}}                                                                          & \multicolumn{4}{c|}{VGGFace2 \cite{21cao2018vggface2}}                                                                     \\ 
\cline{2-9}
                          & \multicolumn{1}{c|}{\multirow{2}{*}{EER, ASR}} & \multicolumn{3}{c|}{FRR, ASR}                        & \multicolumn{1}{c|}{\multirow{2}{*}{EER, ASR}} & \multicolumn{3}{c|}{FRR, ASR}                      \\ 
\cline{3-5}\cline{7-9}
                          & \multicolumn{1}{c|}{}                          & FAR=0.1         & FAR=0.01         & FAR=0.001       & \multicolumn{1}{c|}{}                          & FAR=0.1        & FAR=0.01       & FAR=0.001        \\ 
\hline
(i)                       & 10.67\%, 81.3\%                                & 10.7\%, 79.7\%  & 30.4\%, 42.0\%   & 47.3\%, 24.5\%  & 2.59\%, 81.7\%                                 & 0.7\%, 59.0\%  & 4.0\%, 41.0\%  & 10.5\%, 20.6\%   \\
(ii)                      & 21.85\%, 87.3\%                                & 33.7\%, 64.6\%  & 74.2\%, 12.8\%   & 100.0\%, 0.0\%  & 5.0\%, 71.6\%                                  & 2.9\%, 57.8\%  & 12.9\%, 32.2\% & 35.9\%, 13.5\%   \\
(iii)                     & 29.44\%, 79.8\%                                & 46.23\%, 59.4\% & 69.17\%, 27.16\% & 80.64\%, 11.6\% & 12.77\%, 69.4\%                                & 13.7\%, 66.3\% & 27.4\%, 34.0\% & 47.4\%, 13.85\%  \\
(iv)                      & 3.22\%, 47.5\%                                 & 1.93\%, \_\_    & 9.6\%, 15.6\%    & 41.6\%, 0.2\%   & 2.25\%, 56.3\%                                 & 0.16\%, \_\_   & 4.6\%, 38.9\%  & 25.16\%, 5.2\%   \\
\hline
\end{tabular}%
}
\end{table*}

In terms of ASR, while the highest percentage on CASIA ($87.3\%$) belongs to scenario (ii) at EER point, on VGGFace2 it happens on scenario (i) at EER with $81.7\%$. Regarding the proposed method, the corresponding values for the ASR at the EER point are of $47.5\%$ and $56.3\%$ on CASIA and VGGFace2 respectively.

These results show the superiority of OTB-morph compared to related methods both in security protection and recognition performance.

\section{Conclusions}
This work introduces a new type of cancelable biometrics method, which can be categorized as a branch of visual cryptography with the aim of protecting the biometric templates of clients against all kinds of leakage attacks. We adapted the concept of one-time pad to biometrics by using random biometrics as auxiliary data in a cancelable biometrics scheme called OTB-morph (One-Time Biometrics via Morphing). We then experimented with a practical implementation for face biometrics via face morphing. Regarding the transformation function, a morphing algorithm based on Dlib and OpenCV is used for generating the cancelable templates. The proposed method improves both the biometric performance and security by using a random face morphed with the face of a client in every verification session. Therefore, the client is able to exploit the server's services without revealing his actual face. Since the client face is changing in every session, it is very difficult even for the server to find out the real identity of him. The results taken from implementing the proposed method confirm that not only our method surpass unprotected biometric verification in terms of recognition performance but also it excels reducing the attack success rate compared to other evaluated protection scenarios.

In our future work we will implement different methods in longer iterations, explore the challenges and opportunities for improving the proposed OTB-morph when template update schemes are used for dealing with aging biometrics \cite{galbally13PONEagingSignature,2020_IPAS_AdaptiveFace_Orru}, the application of time-adaptive biometrics \cite{fierrez06phd,10.1145/3344255}, and how to connect OTB-morph with distributed approaches for storing the templates \cite{2019_CVPRw_BioBlockStorage_Delgado}.

\section*{Acknowledgments}
This work has been supported by projects: PRIMA (ITN-2019-860315), TRESPASS-ETN (ITN-2019-860813), and BIBECA (RTI2018-101248-B-I00 MINECO/FEDER). M.G. is supported by PRIMA and I.S. is supported by a FPI fellowship from Univ. Autonoma de Madrid.

{\small
\bibliographystyle{ieee_fullname}
\bibliography{references}
}

\end{document}